\documentclass[11pt,a4paper]{article}
\usepackage[hyperref]{emnlp-ijcnlp-2019}
\usepackage{times}
\usepackage{latexsym}
\usepackage{graphicx}
\usepackage{url}
\usepackage{cleveref}
\usepackage{breakurl}
\usepackage{float}
\restylefloat{table}

\usepackage[switch]{lineno}    
\usepackage{fancyhdr}
\crefformat{section}{\S#2#1#3} 
\crefformat{subsection}{\S#2#1#3}
\crefformat{subsubsection}{\S#2#1#3}

\aclfinalcopy 

\title{Gunrock: A Social Bot for Complex and Engaging Long Conversations}

\author{Dian Yu, Michelle Cohn, Yi Mang Yang, Chun-Yen Chen, Weiming Wen\\ \textbf{Jiaping Zhang, Mingyang Zhou, Kevin Jesse, Austin Chau, Antara Bhowmick}\\ \textbf{ Shreenath Iyer, Giritheja Sreenivasulu, Sam Davidson, Ashwin Bhandare, Zhou Yu} \\
  University of California, Davis \\
  {\{dianyu, mdcohn, yimyang, abtchen, wmwen, jpzhang, minzhou, krjesse\}@ucdavis.edu}\\
  {\{amchau, abhowmick, shriyer, s.giritheja, ssdavidson, asbhandar, joyu\}@ucdavis.edu} }

\date{}

\begin{document}
\maketitle
\begin{abstract}
Gunrock is the winner of the 2018 Amazon Alexa Prize, as evaluated by coherence and engagement from both real users and Amazon-selected expert conversationalists. We focus on understanding complex sentences and having in-depth conversations in open domains. In this paper, we introduce some innovative system designs and related validation analysis. Overall, we found that users produce longer sentences to Gunrock, which are directly related to users' engagement (e.g., ratings, number of turns). Additionally, users' backstory queries about Gunrock are positively correlated to user satisfaction. Finally, we found dialog flows that interleave facts and personal opinions and stories lead to better user satisfaction. 

\end{abstract}

\section{Introduction}
Amazon Alexa Prize \cite{alexa_prize} provides a platform to collect real human-machine conversation data and evaluate performance on speech-based social conversational systems. 
Our system, Gunrock \cite{gunrock}\footnote{\url{https://gunrock-ucdavis.weebly.com/2018-system-demonstration.html}}
addresses several limitations of prior chatbots \cite{context, persona_chat, soundingboard} including inconsistency and difficulty in complex sentence understanding (e.g., long utterances) and 
provides several contributions: 
First, Gunrock's  multi-step language understanding modules enable the system to provide more useful information to the dialog manager, including a novel dialog act scheme. Additionally, the natural language understanding (NLU) module can handle more complex sentences, including those with coreference.
Second, Gunrock interleaves actions to elicit users' opinions and provide responses to create an in-depth, engaging conversation; while a related strategy to interleave task- and non-task functions in chatbots has been proposed \cite{rudnicky2019c}, no chatbots to our knowledge have employed a fact/opinion interleaving strategy. 
Finally, we use an extensive persona database to provide coherent profile information, a critical challenge in building social chatbots \cite{persona_chat}. Compared to previous systems \cite{soundingboard}, Gunrock generates more balanced conversations between human and machine by encouraging and understanding more human inputs (see Table \ref{table:chat_example} for an example).
\begin{table}[H]
\small
\def\arraystretch{0.3}
\begin{tabular}{l p{0.68\linewidth}}
\textbf{User\_1}: & let's chat\\
\textbf{Gunrock\_1}: & Hi, this is an Alexa Prize Socialbot. It's nice to talk to you again, George. I think last time we were talking about movies. Would you like to talk about it now?\\
\textbf{User\_2}: & sure [SPL] that would be great\\
\textbf{Gunrock\_2}: & By the way, I love to discover new movies. I recently watched Crazy Rich Asians because someone told me it was hilarious. What movie would you recommend to me?\\
\textbf{User\_3}: & ha [SPL] it's a tough question [SPL] i don't think i have a good one to recommend [SPL] wait [SPL] i think that \textcolor{red}{\textit{stars born}} is good\\
\textbf{Gunrock\_3}: &  Yes I remember watching \textcolor{green}{\textit{a star is born}}. Hmm. If you had to rate this movie from 1 to 10, what would it be? \\
\textbf{User\_4}: & ten  \\
\textbf{Gunrock\_4}: &  A perfect 10! You have to tell me more. What was so exceptional? When I watched it, the music design really stood out to me.\\
\textbf{User\_5}: & when i watched it the music was amazing [SPL] and bradley cooper was super talented in the movie [SPL] i really like him \\
\textbf{Gunrock\_5}: & You've got me wanting to watch the movie again. Ouu, Maybe you will find this interesting. Bradley Cooper claimed that his dream was to be a director before he got into acting and this was his directorial debut ...
\end{tabular}
\caption{An example conversation between Gunrock and a human user (User). Text in red represents ASR errors while text in green represents ASR correction. [SPL] represents a split of the sentences from a sequence of words from ASR.}
\label{table:chat_example}
\end{table}

\section{System Architecture}
Figure \ref{fig:architecture} provides an overview of Gunrock's architecture. We extend the Amazon Conversational Bot Toolkit (CoBot) \cite{alexa_prize_18} which is a flexible event-driven framework. CoBot provides ASR results and natural language processing pipelines through the Alexa Skills Kit (ASK) \cite{ask}. Gunrock corrects ASR according to the context (\cref{asr}) and creates a natural language understanding (NLU) (\cref{nlu}) module where multiple components analyze the user utterances. A dialog manager (DM) (\cref{dm}) uses features from NLU to select topic dialog modules and defines an individual dialog flow. Each dialog module leverages several knowledge bases (\cref{knowledge}). Then a natural language generation (NLG) (\cref{nlg}) module generates a corresponding response. Finally, we markup the synthesized responses and return to the users through text to speech (TTS) (\cref{tts}). While we provide an overview of the system in the following sections, for detailed system implementation details, please see the technical report \cite{gunrock}.

\begin{figure*}[t!]
\label{fig:architecture}
    \centering
     \includegraphics[width=\textwidth, height=8cm]{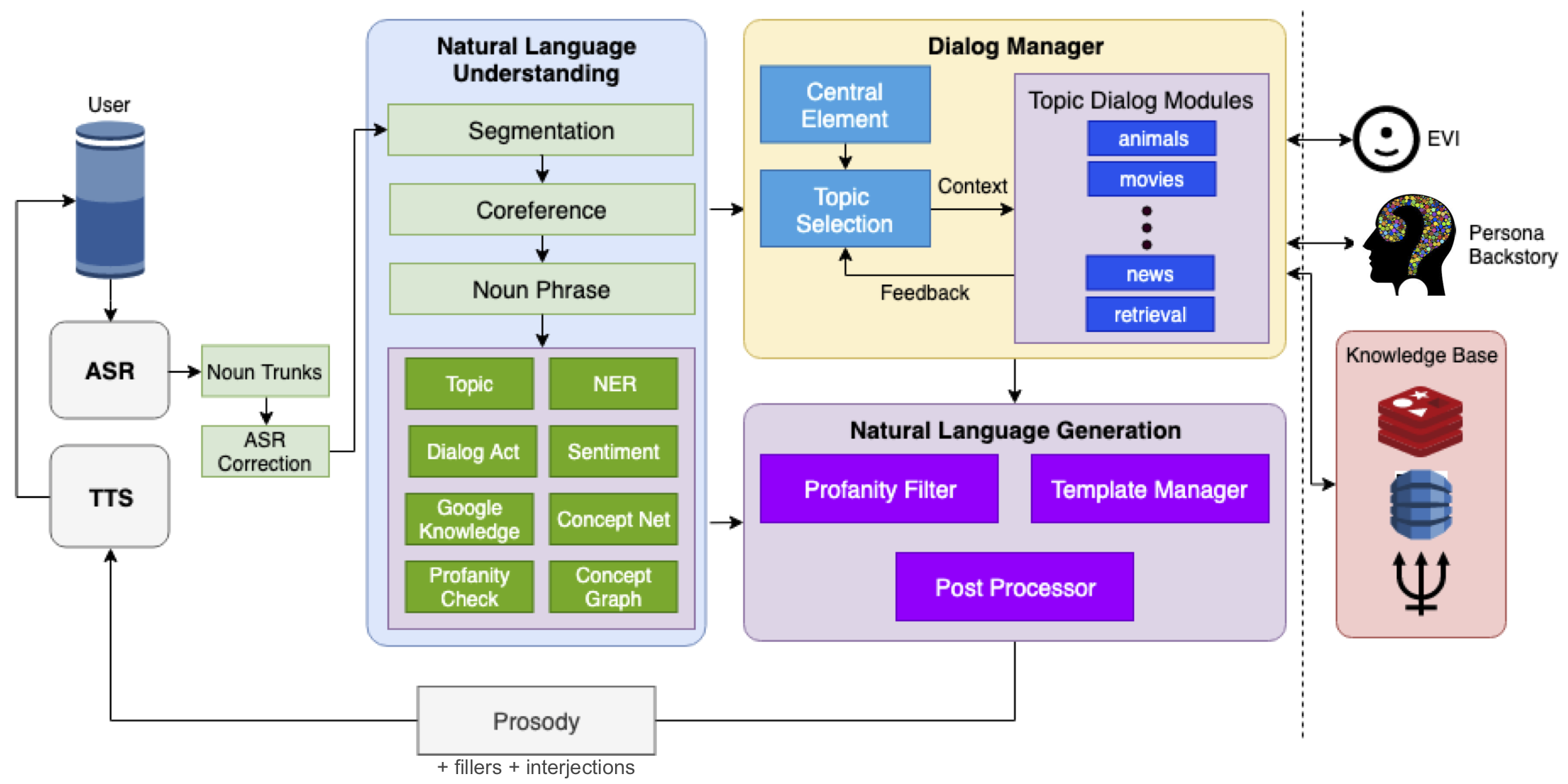}
    \caption{\label{fig:architecture} Gunrock system architecture}
\end{figure*}

\subsection{Automatic Speech Recognition}
\label{asr}
Gunrock receives ASR results with the raw text and timestep information for each word in the sequence (without case information and punctuation). Keywords, especially named entities such as movie names, are prone to generate ASR errors without contextual information, but are essential for NLU and NLG. Therefore, Gunrock uses domain knowledge to correct these errors by comparing noun phrases to a knowledge base (e.g. a list of the most popular movies names) based on their phonetic information. We extract the primary and secondary code using The Double Metaphone Search Algorithm \cite{dmetaphone} for noun phrases (extracted by noun trunks) and the selected knowledge base, and suggest a potential fix by code matching. An example can be seen in User\_3 and Gunrock\_3 in Table \ref{table:chat_example}.


\subsection{Natural Language Understanding}
\label{nlu}
Gunrock is designed to engage users in deeper conversation; accordingly, a user utterance can consist of multiple units with complete semantic meanings. We first split the corrected raw ASR text into sentences by inserting break tokens. 
An example is shown in User\_3  in Table \ref{table:chat_example}. Meanwhile, we mask named entities before segmentation so that a named entity will not be segmented into multiple parts and an utterance with a complete meaning is maintained (e.g.,``i like the movie a star is born"). We also leverage timestep information to filter out false positive corrections. After segmentation, our coreference implementation leverages entity knowledge (such as person versus event) and replaces nouns with their actual reference by entity ranking. 
We implement 
coreference resolution on entities both within segments in a single turn as well as across multiple turns. For instance, ``him" in the last segment in User\_5 is replaced with ``bradley cooper" in Table \ref{table:chat_example}. Next, we use a constituency parser to generate noun phrases from each modified segment. Within the sequence pipeline to generate complete segments, Gunrock detects (1) topic, (2) named entities, and (3) sentiment using ASK in parallel. The NLU module uses knowledge graphs including Google Knowledge Graph \footnote{\url{https://developers.google.com/knowledge-graph/}} to call for a detailed description of each noun phrase for understanding. 

In order to extract the intent for each segment, we designed MIDAS, a human-machine dialog act scheme with 23 tags and implemented a multi-label dialog act classification model using contextual information \cite{MIDAS}. Next, the NLU components analyzed on each segment in a user utterance are sent to the DM and NLG module for state tracking and generation, respectively.

\subsection{Dialog Manager}
\label{dm}
We implemented a hierarchical dialog manager, consisting of a high level and low level DMs. The former leverages NLU outputs for each segment and selects the most important segment for the system as the central element using heuristics. For example, ``i just finished \textit{reading} harry potter," triggers Sub-DM: Books. Utilizing the central element and features extracted from NLU, input utterances are mapped onto 11 possible topic dialog modules (e.g., movies, books, animals, etc.), including a backup module, retrieval. 
\newline \indent  Low level dialog management is handled by the separate topic dialog modules, which use modular finite state transducers to execute various dialog segments processed by the NLU. Using topic-specific modules enables deeper conversations that maintain the context. We design dialog flows in each of the finite state machines, as well. Dialog flow is determined by rule-based transitions between a specified fixed set of dialog states. To ensure that our states and transitions are effective, we leverage large scale user data to find high probability responses and high priority responses to handle in different contexts. Meanwhile, dialog flow is customized to each user by tracking user attributes as dialog context. In addition, each dialog flow is adaptive to user responses to show acknowledgement and understanding (e.g., talking about pet ownership in the animal module). Based on the user responses, many dialog flow variations exist to provide a fresh experience each time. This reduces the feeling of dialogs being scripted and repetitive. Our dialog flows additionally interleave facts, opinions, experiences, and questions to make the conversation flexible and interesting. 
\newline \indent In the meantime, we consider feedback signals such as ``continue" and ``stop" from the current topic dialog module, indicating whether it is able to respond to the following request in the dialog flow, in order to select the best response module. Additionally, in all modules we allow mixed-initiative interactions; users can trigger a new dialog module when they want to switch topics while in any state. For example, users can start a new conversation about movies from any other topic module.



\subsection{Knowledge Databases}
\label{knowledge}
All topic dialog modules query knowledge bases to provide information to the user. To respond to general factual questions, Gunrock queries the EVI factual database \footnote{\url{https://www.evi.com/}}, as well as other up-to-date scraped information appropriate for the submodule, such as news and current showing movies in a specific location from databases including IMDB. One contribution of Gunrock is the extensive Gunrock Persona Backstory database, consisting of over 1,000 responses to possible questions for Gunrock as well as reasoning for her responses for roughly 250 questions (see Table 2). We designed the system responses to elicit a consistent personality within and across modules, modeled as a female individual who is positive, outgoing, and is interested in science and technology. 


\begin{table}[ht!]
\small
\def\arraystretch{1}
\begin{tabular}{l p{0.75\linewidth}}
\textbf{User1}: & do you like jeopardy? \\
\textbf{Gunrock1}: & Sure I like Jeopardy, especially when Watson competed. \\
\textbf{User2}: & why? \\
\textbf{Gunrock2}: & I'm so impressed with the capabilities of a supercomputer. \\
\end{tabular}
\caption{Example interaction between Gunrock and a human user (User) querying Gunrock's backstory.}
\end{table}

\subsection{Natural Language Generation}
\label{nlg}
In order to avoid repetitive and non-specific responses commonly seen in dialog systems \cite{non-specific}, Gunrock uses a template manager to select from a handcrafted response templates based on the dialog state. One dialog state can map to multiple response templates with similar semantic or functional content but differing surface forms. Among these response templates for the same dialog state, one is randomly selected without repetition to provide variety unless all have been exhausted. When a response template is selected, any slots are substituted with actual contents, including queried information for news and specific data for weather. For example, to ground a movie name due to ASR errors or multiple versions, one template is ``Are you talking about \{movie\_title\} released in \{release\_year\} starring \{actor\_name\} as \{actor\_role\}?". Module-specific templates were generated for each topic (e.g., animals), but some of the templates are generalizable across different modules (e.g., ``What’s your favorite [movie $|$ book $|$ place to visit]?") 
\newline \indent In many cases, response templates corresponding to different dialog acts are dynamically composed to give the final response. For example, an appropriate acknowledgement for the user’s response can be combined with a predetermined follow-up question.  

\subsection{Text To Speech}
\label{tts}
After NLG, we adjust the TTS of the system to improve the expressiveness of the voice to convey that the system is an engaged and active participant in the conversation. We use a rule-based system to systematically add interjections, specifically Alexa Speechcons, 
and fillers to approximate human-like cognitive-emotional expression \cite{Tokuhisa:2009:RUE:1654595.1654628}. For more on the framework and analysis of the TTS modifications, see \cite{speechcon_paper}.




\section{Analysis}
From January 5, 2019 to March 5, 2019, we collected conversational data for Gunrock.  During this time, no other code updates occurred. 
We analyzed conversations for Gunrock with at least 3 user turns to avoid conversations triggered by accident. Overall, this resulted in a total of 34,432 user conversations. Together, these users gave Gunrock an average rating of 3.65 (median: 4.0), which was elicited at the end of the conversation (``On a scale from 1 to 5 stars, how do you feel about talking to this socialbot again?"). Users engaged with Gunrock for an average of 20.92 overall turns (median 13.0), with an average of 6.98 words per utterance, and had an average conversation time of 7.33 minutes (median: 2.87 min.). We conducted three principal analyses: users' response depth (\cref{word_count}), backstory queries (\cref{backstory_persona}), and interleaving of personal and factual responses (\cref{pets}).

\subsection{Response Depth: Mean Word Count} 
\label{word_count}
Two unique features of Gunrock are its ability to dissect longer, complex sentences, and its methods to encourage users to be active conversationalists, elaborating on their responses. In prior work, even if users are able to drive the conversation, often bots use simple yes/no questions to control the conversational flow to improve understanding; as a result, users are more passive interlocutors in the conversation. We aimed to improve user engagement by designing the conversation to have more open-ended opinion/personal questions, and show that the system can understand the users' complex utterances (See \cref{nlu} for details on NLU). Accordingly, we ask if users' speech behavior will reflect Gunrock's technical capability and conversational strategy, producing longer sentences.

We assessed the degree of conversational depth by measuring users' mean word count. Prior work has found that an increase in word count has been linked to improved user engagement (e.g., in a social dialog system \cite{zhou_diss}). For each user conversation, we extracted the overall rating, the number of turns of the interaction, and the user's per-utterance word count (averaged across all utterances). We modeled the relationship between word count and the two metrics of user engagement (overall rating, mean number of turns) in separate linear regressions. 
\begin{figure}[!ht]
    \centering
     \includegraphics[height=5.5cm, width=6cm]{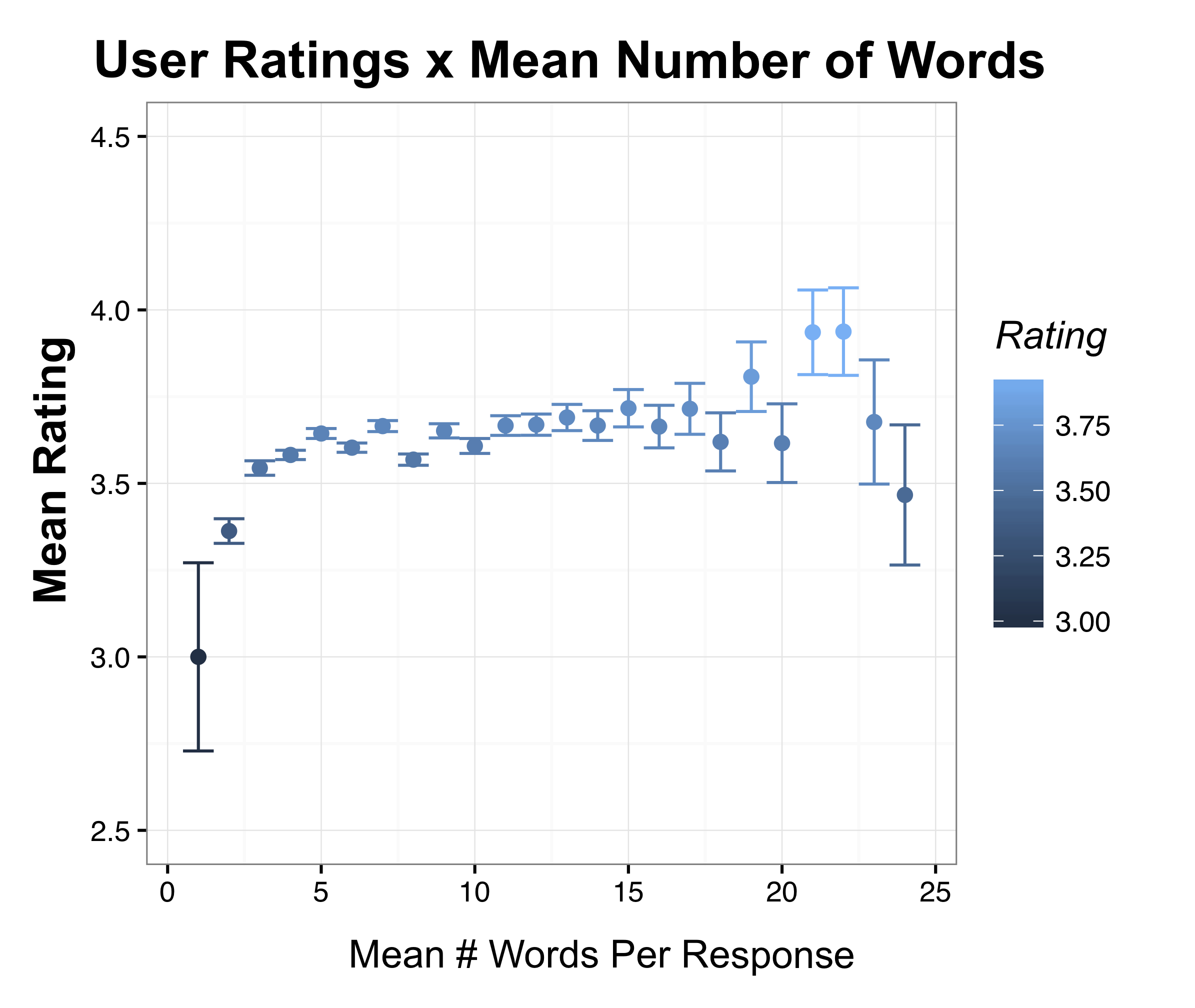}
    \caption{\label{fig:scheme1} Mean user rating by mean number of words. Error bars show standard error.}
\end{figure}

Results showed that users who, on average, produced utterances with more words gave significantly higher ratings ($\beta$=0.01, \textit{SE}=0.002, \textit{t}=4.79, \textit{p}$<$0.001)\footnote{$\beta$ = Unstandardized Beta Coefficient, \textit{SE} = Standard Error, \textit{t} = T value, \textit{p} = P value}(see Figure 2) and engaged with Gunrock for significantly greater number of turns ($\beta$=1.85, \textit{SE}=0.05, \textit{t}=35.58, \textit{p}$<$0.001) (see Figure 2). These results can be interpreted as evidence for Gunrock's ability to handle complex sentences, where users are not constrained to simple responses to be understood and feel engaged in the conversation -- and evidence that individuals are more satisfied with the conversation when they take a more active role, rather than the system dominating the dialog. On the other hand, another interpretation is that users who are more talkative may enjoy talking to the bot in general, and thus give higher ratings in tandem with higher average word counts.
\subsection{Gunrock's Backstory and Persona}
\label{backstory_persona}
We assessed the user's interest in Gunrock by tagging instances where the user triggered Gunrock's backstory (e.g., ``What's your favorite color?"). For users with at least one backstory question, we modeled overall (log) Rating with a linear regression by the (log) `Number of Backstory Questions Asked' (log transformed due to the variables' nonlinear relationship). We hypothesized that users who show greater curiosity about Gunrock will display higher overall ratings for the conversation based on her responses. Overall, the number of times users queried Gunrock's backstory was strongly related to the rating they gave at the end of the interaction  (log:$\beta$=0.10, \textit{SE}=0.002, \textit{t}=58.4, \textit{p}$<$0.001)(see Figure 3). This suggests that maintaining a consistent personality --- and having enough responses to questions the users are interested in --- may improve user satisfaction. 

\begin{figure}[h!]
    \centering
     \includegraphics[height=5.5cm, width=6.5cm]{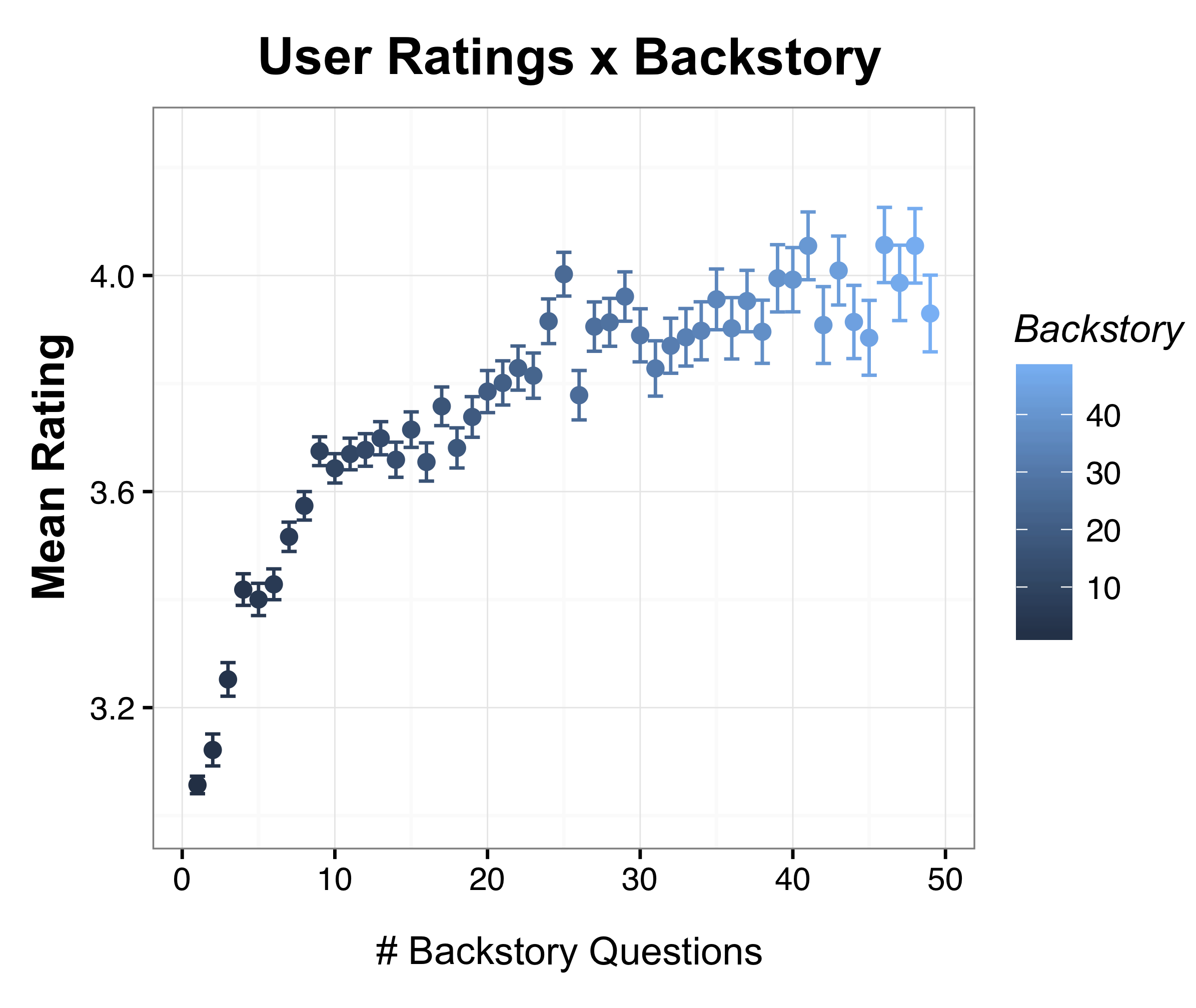}
    \caption{\label{fig:scheme1} Mean user rating based on number of queries to Gunrock's backstory. Error bars show standard error.}
\end{figure}

\subsection{Interleaving Personal and Factual Information: Animal Module}
\label{pets}
Gunrock includes a specific topic module on animals, which includes a factual component where the system provides animal facts, as well as a more personalized component about pets. Our system is designed to engage users about animals in a more casual conversational style \cite{ventola1979structure}, eliciting follow-up questions if the user indicates they have a pet; if we are able to extract the pet's name, we refer to it in the conversation (e.g., ``Oliver is a great name for a cat!", ``How long have you had Oliver?"). In cases where the user does not indicate that they have a pet, the system solely provides animal facts. Therefore, the animal module can serve as a test of our interleaving strategy: we hypothesized that combining facts and personal questions --- in this case about the user's pet --- would lead to greater user satisfaction overall. 



\indent We extracted conversations where Gunrock asked the user if they had ever had a pet and categorized responses as ``Yes", ``No", or ``NA" (if users did not respond with an affirmative or negative response). 
We modeled user rating with a linear regression model, with predictor of ``Has Pet' (2 levels: Yes, No). We found that users who talked to Gunrock about their pet showed significantly higher overall ratings of the conversation ($\beta$=0.15, \textit{SE}=0.06, \textit{t}=2.53, \textit{p}$=$0.016) (see Figure 4). One interpretation is that interleaving factual information with more in-depth questions about their pet result in improved user experience. Yet, another interpretation is that pet owners may be more friendly and amenable to a socialbot; for example, prior research has linked differences in personality to pet ownership \cite{kidd1980personality}.

\begin{figure}[!ht]
    \centering
     \includegraphics[height=4.5cm, width=5cm]{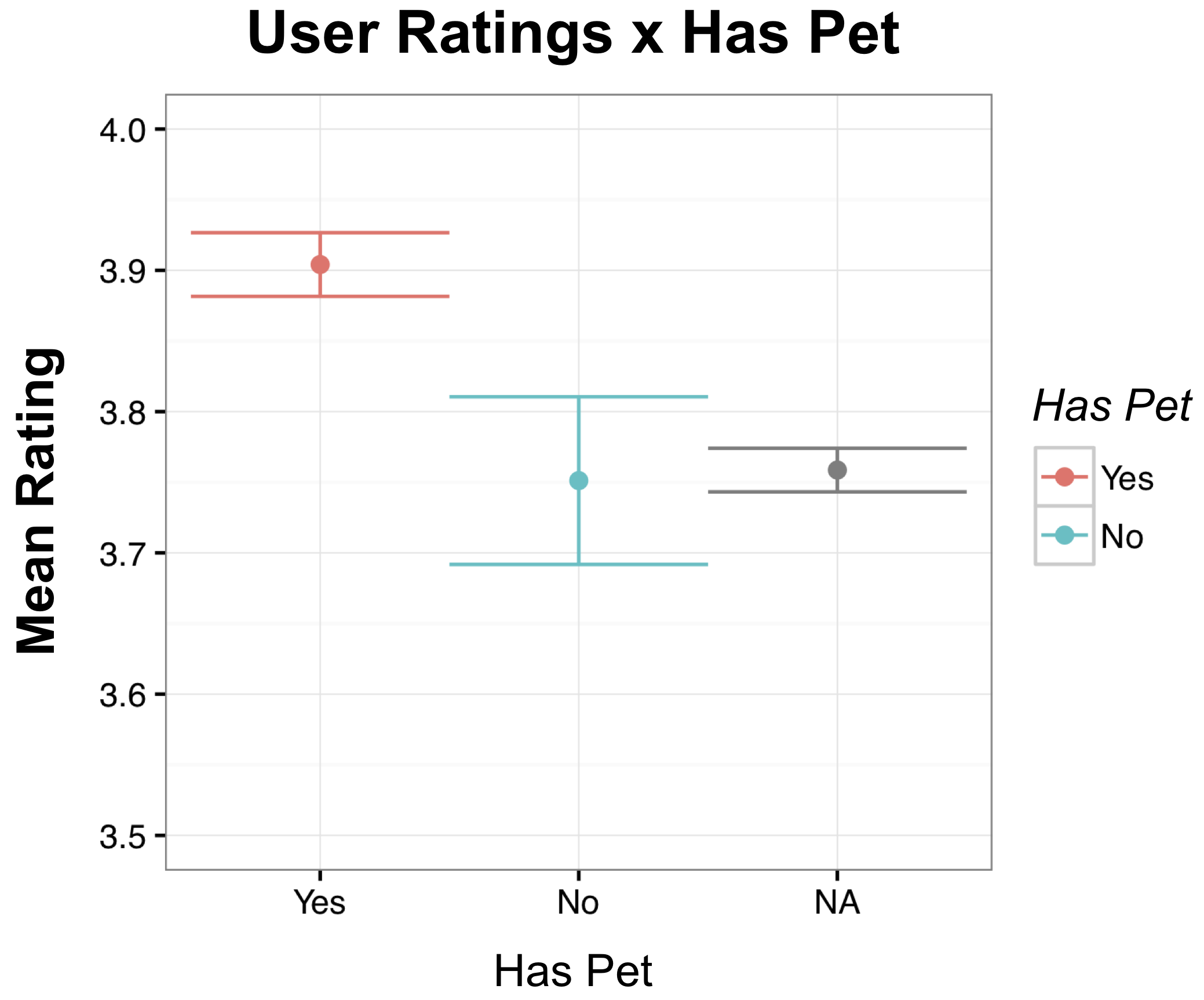}
    \caption{\label{fig:scheme1} Mean user rating based 'Has Pet'. Error bars show standard error.}
\end{figure}

 


\section{Conclusion}
Gunrock is a social chatbot that focuses on having long and engaging speech-based conversations with thousands of real users. Accordingly, our architecture employs specific modules to handle longer and complex utterances and encourages users to be more active in a conversation. Analysis shows that users' speech behavior reflects these capabilities. Longer sentences and more questions about Gunrocks's backstory positively correlate with user experience. Additionally, we find evidence for interleaved dialog flow, where combining factual information with personal opinions and stories improve user satisfaction. 
Overall, this work has practical applications, in applying these design principles to other social chatbots, as well as theoretical implications, in terms of the nature of human-computer interaction (cf. 'Computers are Social Actors' \cite{nass1994computers}). Our results suggest that users are engaging with Gunrock in similar ways to other humans: in chitchat about general topics (e.g., animals, movies, etc.), taking interest in Gunrock's backstory and persona, and even producing more information about themselves in return. 




 




\section*{Acknowledgments}
We would like to acknowledge the help from Amazon in terms of financial and technical support. 

\bibliography{emnlp-ijcnlp-2019}
\bibliographystyle{acl_natbib}

\appendix

\end{document}